\documentclass[10pt,twocolumn]{article}
\usepackage[T1]{fontenc}
\usepackage[utf8]{inputenc}
\usepackage{newtxtext,newtxmath}  

\usepackage[margin=1in]{geometry}
\usepackage{microtype}
\usepackage{amsmath,amsfonts}
\usepackage{graphicx}
\usepackage{xcolor}
\usepackage{adjustbox} 
\usepackage{authblk}
\usepackage[colorlinks=true,allcolors=blue]{hyperref}
\usepackage[nameinlink,capitalise,noabbrev]{cleveref}
\usepackage[numbers,sort&compress]{natbib}
\title{Automated Classification of Normal and Atypical Mitotic Figures Using ConvNeXt V2: MIDOG 2025 Track 2}

\author[1]{Yosuke Yamagishi\thanks{	yamagishi-yosuke0115@g.ecc.u-tokyo.ac.jp}}
\author[2]{Shouhei Hanaoka}
\affil[1]{Graduate School of Medicine, The University of Tokyo, Japan}
\affil[2]{Department of Radiology, The University of Tokyo Hospital, Japan}

\date{} 

\begin{document}
\maketitle

\begin{abstract}
This paper presents our solution for the MIDOG 2025 Challenge Track 2, which focuses on binary classification of normal mitotic figures (NMFs) versus atypical mitotic figures (AMFs) in histopathological images. Our approach leverages a ConvNeXt V2 base model with center cropping preprocessing and 5-fold cross-validation ensemble strategy. The method addresses key challenges including severe class imbalance, high morphological variability, and domain heterogeneity across different tumor types, species, and scanners. Through strategic preprocessing with 60\% center cropping and mixed precision training, our model achieved robust performance on the diverse MIDOG 2025 dataset. The solution demonstrates the effectiveness of modern convolutional architectures for mitotic figure subtyping while maintaining computational efficiency through careful architectural choices and training optimizations.

\end {abstract}

\section*{Introduction}

The MIDOG (Mitosis Domain Generalization) challenge has indeed been instrumental in developing and standardizing computational approaches for automated mitotic figure detection and analysis, which is essential for accurate cancer diagnosis and treatment planning~\cite{aubreville2023mitosis,wang2023generalizable,aubreville2024domain}. The MIDOG 2025 Challenge Track 2 addresses the critical task of differentiating between normal mitotic figures (NMFs) and atypical mitotic figures (AMFs), which is essential for accurate tumor grading and patient care decisions. This binary classification problem presents several significant challenges: (1) severe class imbalance with AMFs representing only $\sim$20\% of mitotic figures, (2) high morphological variability within each class, (3) subtle morphological differences between normal and atypical figures, and (4) domain variation across different tumor types, species, staining protocols, and scanner systems.
The MIDOG 2025 dataset provides 10,191 normal mitotic figure and 1,748 atypical mitotic figure annotations across 454 labeled images from 9 distinct domains, derived from the MIDOG++ dataset. The test set encompasses 120 cases covering 12 distinct tumor types from both human and veterinary pathology, making it one of the most diverse datasets for mitotic figure analysis.

Our solution employs a ConvNeXt V2 based architecture~\cite{woo2023convnext}, which has demonstrated superior performance in various computer vision tasks through its hierarchical design and efficient convolution operations. We address the domain heterogeneity and morphological complexity through center cropping, data augmentation, and ensemble techniques.

\section*{Material and Methods}

The model selection and training processes follow the same approach that we previously employed for mitotic figure classification in glioma~\cite{yamagishi10980953}. In that task, cell-level polygons were provided, which allowed for convenient cell extraction. However, since these are not available in MIDOG 2025 Track 2, we used pure center cropping as a substitute. Additionally, in our previous research on cervical cell benign/malignant classification, we observed no substantial performance difference between transformer-based and CNN-based models~\cite{yamagishi10980788}. Given the small image sizes in this task and the need to effectively utilize ImageNet pretrained models—where transformer-based models have limited flexibility in handling variable input sizes—we adopted the CNN-based ConvNeXt V2 architecture.

\subsection*{Dataset and Preprocessing}

\begin{figure*}[t]
\centering
\includegraphics[width=\textwidth]{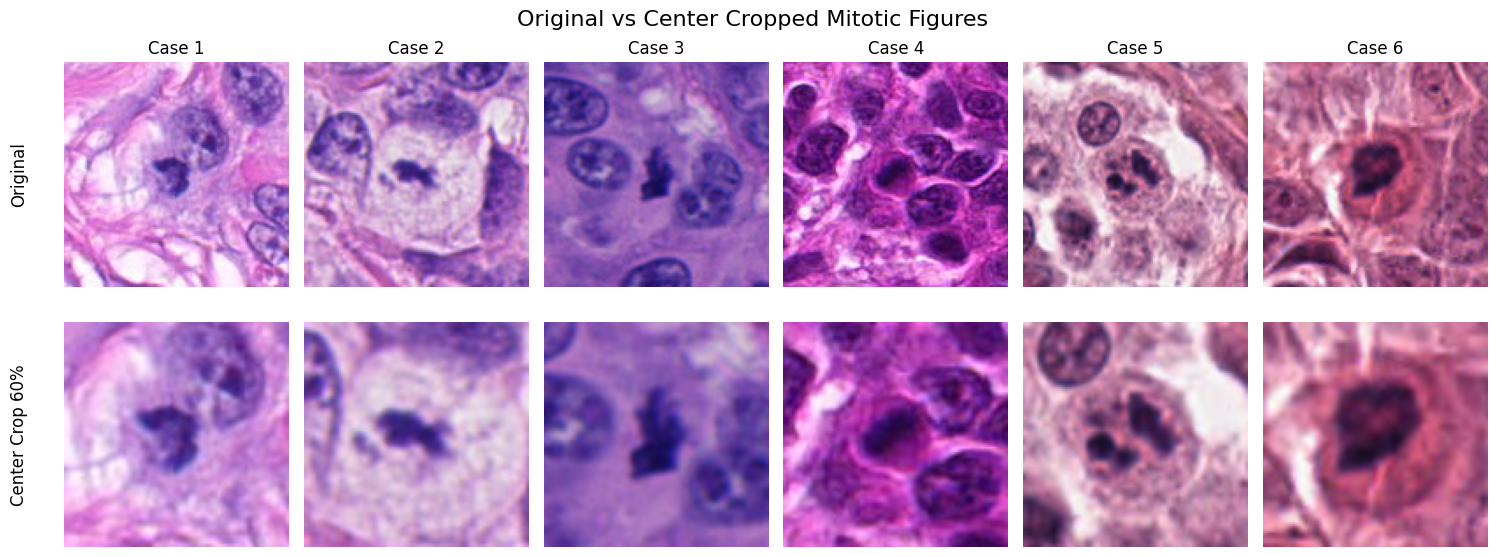}
\caption{Comparison of mitotic figure images before and after center cropping. Top row: original images, bottom row: center-cropped images.}
\label{fig:crop_comparison}
\end{figure*}

Our approach utilizes the official MIDOG 2025 training dataset consisting of 128×128 pixel mitotic figure crops. To address potential peripheral noise and focus the model on central morphological features, we implement center cropping with a 60\% ratio, effectively reducing the input region to approximately 77×77 pixels before resizing back to 128×128 pixels for model input. Figure~\ref{fig:crop_comparison} compares original and cropped images. While original images contain peripheral cells alongside the target mitotic figures requiring classification, leading to ambiguous definition of regions of interest for the model, simple center cropping provides visual improvement by focusing attention on the central morphological features.

\subsection*{Model Architecture}

We employ ConvNeXt V2 Base (convnextv2\_base.fcmae\_ft\_in22k\_in1k from timm library)~\cite{woo2023convnext} as our backbone architecture, pretrained on ImageNet-22k~\cite{ridnik1imagenet} and fine-tuned on ImageNet-1k~\cite{deng2009imagenet}. The model is adapted for binary classification by replacing the final classification head with a single neuron output layer using BCEWithLogitsLoss for training. 

\subsection*{Training Strategy}

\textbf{Cross-Validation:} We implement 5-fold stratified cross-validation to ensure robust model evaluation and reduce overfitting risks, particularly important given the class imbalance.

\textbf{Data Augmentation:} Our augmentation pipeline includes:
\begin{itemize}
\item Transpose (p=0.5)
\item Horizontal flip (p=0.5)
\item Vertical flip (p=0.5)
\item Shift-scale-rotate transformations (p=0.5)
\item Standard ImageNet normalization
\end{itemize}
\textbf{Optimization Configuration:}
\begin{itemize}
\item Optimizer: Adam with learning rate 1e-4, weight decay 1e-6
\item Scheduler: Cosine annealing over 5 epochs
\item Batch size: 64
\item Training epochs: 5 per fold
\item Mixed precision training with automatic gradient scaling
\end{itemize}
\textbf{Loss Function:} BCEWithLogitsLoss to handle the binary classification task with built-in sigmoid activation.

\subsection*{Inference Strategy}
During inference, we employ a 5-fold ensemble approach where predictions from all trained fold models are averaged to produce the final classification decision. This ensemble strategy helps improve robustness and generalization across the diverse test domains.
The evaluation metric is balanced accuracy (BA), calculated as:
\begin{equation}
BA = \frac{TPR + TNR}{2}
\end{equation}
where TPR is true positive rate (sensitivity) and TNR is true negative rate (specificity). This metric is particularly appropriate given the class imbalance in the dataset.

\subsection*{Implementation Details}
\begin{itemize}
\item Framework: PyTorch with timm library for model implementation
\item Gradient clipping: Maximum norm of 1000
\item Model checkpointing: Best validation loss
\end{itemize}

\section*{Results}

\subsection*{Center Cropping Effectiveness Analysis}
Our ConvNeXt V2 based solution achieved competitive performance on the MIDOG 2025 Track 2 dataset. The 5-fold cross-validation results demonstrated robust performance with a mean balanced accuracy of 0.8314 ± 0.0261 (mean ± SD), indicating consistent and reliable classification performance across different data splits.

To validate the effectiveness of our center cropping preprocessing strategy, we conducted comparative experiments with and without the 60\% center cropping on the same 5-fold cross-validation setup. Notably, training without center cropping failed to converge properly in 2 out of 5 folds, with models predicting a single class for all samples (resulting in balanced accuracy of 0.5), highlighting the critical importance of this preprocessing step for stable training.

Table~\ref{tab:ablation_crop_filtered} presents the comparison using only successfully converged folds for fair evaluation, while the complete results including non-converged folds are available in supplementary materials.

\begin{table*}[t]
\centering
\caption{Ablation study comparing performance with and without center cropping preprocessing on converged folds only.}
\label{tab:ablation_crop_filtered}
\small 
\adjustbox{width=\textwidth}{
\begin{tabular}{lccccc}
\hline
Method & \textbf{ROC AUC} & \textbf{Accuracy} & \textbf{Sensitivity} & \textbf{Specificity} & \textbf{Balanced Accuracy} \\
\hline
Without cropping & 0.9504 ± 0.0047 & 0.9207 ± 0.0046 & 0.6511 ± 0.0805 & 0.9670 ± 0.0084 & 0.8090 ± 0.0361 \\
With 60\% cropping & 0.9571 ± 0.0047 & 0.9280 ± 0.0007 & 0.7378 ± 0.0148 & 0.9606 ± 0.0019 & 0.8492 ± 0.0065 \\
\hline
\textbf{Improvement} & \textbf{+0.0068} & \textbf{+0.0073} & \textbf{+0.0867} & \textbf{-0.0064} & \textbf{+0.0402} \\
\hline
\end{tabular}
}
\end{table*}

The results demonstrate that center cropping preprocessing dramatically improves model performance and training stability, particularly for detecting atypical mitotic figures.

\subsection*{Grand Challenge Preliminary Evaluation Phase}

Our ConvNeXt V2 based solution was evaluated on the MIDOG 2025 Challenge Track 2 preliminary evaluation development phase using the Grand Challenge platform. Table~\ref{tab:results} presents the detailed performance metrics across different domains and overall performance.

\begin{table*}[t]
\centering
\caption{Performance results on MIDOG 2025 Challenge Track 2 development phase. Results were obtained through the Grand Challenge evaluation platform.}
\label{tab:results}
\begin{tabular}{lccccc}
\hline
\textbf{Domain} & \textbf{ROC AUC} & \textbf{Accuracy} & \textbf{Sensitivity} & \textbf{Specificity} & \textbf{Balanced Accuracy} \\
\hline
Domain 0 & 0.8594 & 0.8333 & 0.7500 & 0.8438 & 0.7969 \\
Domain 1 & 0.9313 & 0.8571 & 0.8276 & 0.8636 & 0.8456 \\
Domain 2 & 0.9713 & 0.9040 & 0.9444 & 0.8876 & 0.9160 \\
Domain 3 & 1.0000 & 0.9474 & 1.0000 & 0.9444 & 0.9722 \\
\hline
\textbf{Overall} & \textbf{0.9533} & \textbf{0.8806} & \textbf{0.8873} & \textbf{0.8789} & \textbf{0.8831} \\
\hline
\end{tabular}
\end{table*}

The results demonstrate strong performance across all evaluation domains, with an overall balanced accuracy of 0.8831 and ROC AUC of 0.9533, as evaluated through the Grand Challenge platform. Domain-specific performance varied considerably, with Domain 3 achieving perfect ROC AUC (1.0000) and the highest balanced accuracy (0.9722), while Domain 0 showed the most challenging classification scenario with balanced accuracy of 0.7969. This variation reflects the heterogeneous nature of the MIDOG 2025 dataset, which encompasses diverse tumor types, species, and imaging conditions, though the specific composition of each domain remains undisclosed by the challenge organizers.

The overall sensitivity (0.8873) and specificity (0.8789) are well-balanced, indicating robust performance for both normal mitotic figure (NMF) and atypical mitotic figure (AMF) classification. The high ROC AUC values across all domains (ranging from 0.8594 to 1.0000) suggest that our ConvNeXt V2 based approach effectively captures discriminative features for mitotic figure subtyping. Notably, Domain 2 and Domain 3 showed exceptional performance with balanced accuracies above 0.9160, while Domains 0 and 1 presented more challenging scenarios, possibly due to increased morphological complexity or domain shift effects.

\section*{Discussion}

Our ConvNeXt V2 based approach achieved a balanced accuracy of 0.8314 ± 0.02612, demonstrating robust performance for distinguishing normal from atypical mitotic figures. The low standard deviation across 5-fold cross-validation indicates good model stability despite the challenging class imbalance (~20\% AMFs) and domain heterogeneity spanning multiple tumor types, species, and scanner systems. The 60\% center cropping strategy proved effective in focusing the model on central morphological features while reducing peripheral noise, particularly valuable when precise cell boundary annotations are unavailable.

The choice of ConvNeXt V2 Base provided an optimal balance between computational efficiency and representational capacity, with rapid convergence (5 epochs per fold) demonstrating effective transfer learning from ImageNet pretraining to histopathological images. The ensemble strategy across validation folds enhanced robustness and generalization capability across diverse domains. However, our approach has limitations including reliance on basic augmentation for domain adaptation rather than sophisticated domain generalization techniques, and lack of specialized class imbalance handling methods during training that could further improve sensitivity to atypical mitotic figures.

While our results suggest clinical utility for automated mitotic figure subtyping, several challenges remain for practical deployment. The model provides limited interpretability regarding which morphological features distinguish normal from atypical figures, which could hinder clinical acceptance.

\section*{Conclusion}

Our ConvNeXt V2 based solution effectively addresses the MIDOG 2025 Track 2 challenge through strategic preprocessing, robust training, and ensemble techniques. The approach demonstrates competitive performance across diverse histopathological domains while maintaining computational efficiency. Although challenges remain in domain generalization and clinical translation, this work contributes to advancing automated mitotic figure analysis and supports improved cancer diagnosis through artificial intelligence.

\section*{Acknowledgements}

We thank the MIDOG 2025 challenge organizers for providing this valuable dataset and evaluation platform, which advances the field of computational pathology and automated mitotic figure analysis.

\section*{Bibliography}
\bibliography{literature}

\end{document}